\documentclass{vldb}
\usepackage{graphicx}
\usepackage{balance}  

\vldbTitle{
Model Slicing for Supporting Complex Analytics with Elastic Inference Cost and Resource Constraints
}
\vldbAuthors{Shaofeng Cai, Gang Chen, Beng Chin Ooi, Jinyang Gao}
\vldbDOI{https://doi.org/10.14778/3364324.3364325}
\vldbVolume{13}
\vldbNumber{2}
\vldbYear{2019}

\usepackage{hyperref}

\usepackage{twoopt}
\newcommandtwoopt{\TopicWord}[2][m][s]{#1odel #2licing}

\usepackage{subfig}
\usepackage{multirow}
\usepackage{pifont}
\usepackage{tikz}
\usepackage{tabularx}
\usepackage{bbm}

\usepackage[ruled]{algorithm2e} 
\SetKw{KwFrom}{from}

\usepackage{amsmath}

\makeatletter
\newcommand{\thickhline}{%
    \noalign {\ifnum 0=`}\fi \hrule height 1pt
    \futurelet \reserved@a \@xhline
}
\usepackage{multirow}

\begin{document}
\title{\TopicWord[M][S] for Supporting Complex Analytics with Elastic Inference Cost and Resource Constraints}

\numberofauthors{4}
\author{
{Shaofeng Cai$^\dag$, Gang Chen$^\S$, Beng Chin Ooi$^\dag$}, Jinyang Gao$^\ddag$\vspace{1mm}\\
\fontsize{10}{10}\selectfont\itshape \hspace{0mm}
$^\dag$National University of Singapore \hspace{18mm}
\fontsize{10}{10}\selectfont\itshape \hspace{-10mm} 
$^\S$Zhejiang University \hspace{18mm}
$^\ddag$Alibaba Group \\
\fontsize{9}{9}\selectfont\ttfamily\upshape\vspace{-0.0em}\hspace{10mm}
\{shaofeng, ooibc\}@comp.nus.edu.sg \hspace{12mm}
cg@zju.edu.cn \hspace{12mm}
jinyang.gjy@alibaba-inc.com\\ 
}


\maketitle
\begin{abstract}

Deep learning models have been used to support analytics beyond simple aggregation, where
deeper and wider models have been shown to yield great results.
These models consume a huge amount of memory and computational operations.
However, most of the large-scale industrial applications are often computational budget constrained.
In practice, the peak workload of inference service could be 10x higher than the average cases, with the presence of unpredictable extreme cases.
Lots of computational resources could be wasted during off-peak hours and the system may crash when the workload exceeds system capacity.
How to support deep learning services with a dynamic workload cost-efficiently remains a challenging problem. 
In this paper, we address the challenge with a general and novel training scheme called \textit{\TopicWord}, which enables deep learning models to provide predictions within the prescribed computational resource budget dynamically.
\textit{\TopicWord[M][s]} could be viewed as an elastic computation solution without requiring more computational resources.
Succinctly, each layer in the model is divided into \textit{groups} of a contiguous block of basic components (i.e. neurons in dense layers and channels in convolutional layers), and then partially ordered relation is introduced to these groups by enforcing that groups involved in each forward pass always starts from the \textit{first} group to the \textit{dynamically-determined rightmost} group.
Trained by dynamically indexing the rightmost group with a single parameter \textit{slice rate}, the network is engendered to build up group-wise and residual representation. 
Then during inference, a sub-model with fewer groups can be readily deployed for efficiency whose computation is roughly quadratic to the width controlled by the \textit{slice rate}.
Extensive experiments show that models trained with \textit{\TopicWord} can effectively support on-demand workload with elastic inference cost.

\end{abstract}

\section{Introduction}
\label{sec:intro}

Database management systems (DBMS) have been widely used and optimized to support OLAP-style analytics.
In present-day applications, more and more data-driven machine learning based analytics have been grafted into DBMS to support complex analysis (e.g., stock prediction, disease progression analysis) and/or to enable predictive query and system optimization.
To better understand the data and decipher the information that truly counts in the era of Big Data with its ever-increasing data size and complexity, many advanced large-scale machine learning models have been devised, from million-dimension linear models (e.g., Logistic Regression~\cite{richardson2007predicting}, feature selection~\cite{zhang2016materialization}) to complex models like Deep Neural Networks ~\cite{krizhevsky2012imagenet}.
To meet the demand for more complex analytic queries, OLAP database vendors have integrated Machine Learning (ML) libraries into their systems (e.g., SQL Server pymssql\footnote{https://docs.microsoft.com/en-us/sql/connect/python/pymssql/python-sql-driver-pymssql}, DB2 python\_ibm\_db\footnote{https://github.com/ibmdb/python-ibmdb} and etc).
It is widely recognized that the integration of ML analytics into data systems yields seamless effects since the ML task is treated as an operator of the query plan instead of an individual black-box system on top of data systems.
Naturally, a higher-level abstraction provides more space for optimization.
For example, query planning~\cite{pangzi14,msms}, lazy evaluation~\cite{lazy10}, materialization~\cite{zhang2016materialization} and operator optimization~\cite{boehm2016systemml} could be considered in a fine-grained manner.

Cost and accuracy are always the two most crucial criteria considered for analytic tasks.
Lots of research on approximate query processing have been conducted~\cite{li2016wander,bolin2017aqp} to provide faster yet approximate analytical query results in modern large-scale analytical database systems, while such a trade-off is not equally well researched for modern ML analytic tasks, particularly deep neural network models.
There are two characteristics of the inference cost of analytic tasks for deep neural network models.
Firstly, with the development of high-end hardware and large-scale datasets, recent deep models are growing deeper \cite{krizhevsky2012imagenet,he2016deep} and wider \cite{zagoruyko2016wide,xie2017aggregated}.
State-of-the-art models have been designed with up to hundreds of layers and tens of millions of parameters, which leads to a dramatic increase in the inference cost.
For instance, a 152-layer ResNet~\cite{he2016deep} with over 60 million parameters requires up to 20 Giga FLOPs for the inference of one single $224\times224$ image.
The surging computational cost severely affects the viability of many deep models in industry-scale applications.
Secondly, for most of the analytic tasks, the workload is usually not constant, e.g., the number of images per query for person re-id~\cite{zheng2015scalable} service in peak hours could be five times more than the workload in the off-peak hours.
Therefore, such a trade-off should be naturally supported in the inference phase rather than the training phase: using one single deep model with fixed inference cost to support the peak workload could lead to huge amounts of resources wasting in off-peak hours, and may not be able to handle the unexpected extreme workload.
How to trade off the accuracy and cost during deep model inference remains a challenging problem of great importance.

Existing model architecture re-design~\cite{iandola2016squeezenet,howard2017mobilenets} or model compression~\cite{han2015deep,han2015learning,liu2017learning} methods are not able to handle elastic inference satisfactorily, and we shall use an application example to highlight the challenges.
Singles$'$ Day shopping festival\footnote{\href{https://en.wikipedia.org/wiki/Singles\%27\_Day}{https://en.wikipedia.org/wiki/Singles\%27\_Day}} around 11 November was introduced by Taobao.com and is now becoming one of the biggest online shopping festivals around the world. 
In 2018, the Singles$'$ Day festival generated close to 30 billion dollars of sales in one single day and had attracted hundreds of millions of users from more than 200 different countries. The peak level of trade rate reached 0.256 million per second, and 42 million processing in the database in the first half hour. 
In Singles$'$ Day, the search traffic of the e-commerce search engine increases about three times than in a common day, and could be 10x in its first hour.
Meanwhile, the workload of most other services in Alibaba such as OLTP transactions may also hit the peak at the same time~\cite{cao2018tcprt}, and consequently, it is not possible to scale up the service by acquiring more hardware resources from Alibaba Cloud.
The system degradation is often executed in two simple and naive approaches: First, some costly deep learning models are replaced by simple GBDT~\cite{chen2016xgboost,ke2017lightgbm} models; 
Second, the size of the candidate items for ranking is reduced.
The search accuracy suffers dramatically due to the system degradation in such a coarse-grained manner.
With a deep learning model supporting elastic inference cost, the system degradation management can become more fine-grained where the inference cost and accuracy trade-off per query sample can be dynamically determined based on the current system workload.

In this paper, instead of constructing small models based on each individual workload requirement, we propose and address a related but slightly different research problem:
developing a general framework to support deep learning models with elastic inference cost. 
We base the framework on a pay-as-you-go model to support dynamic trade-offs between computation cost and accuracy during inference time.
That is, dynamic optimization is supported based on system workload, availability of resources and user requirements.
An ML model abstraction with elastic inference cost would greatly benefit the optimization of the system design for complex analytics.
We shall examine the problem from a fresh system perspective and propose our solution -- \textit{\TopicWord}, a general network training mechanism supporting elastics inference cost, to satisfy the run-time memory and computation budget dynamically during the inference phase.
The crux of our approach is to decompose each layer of the model into groups of a contiguous block of basic components, i.e. neurons in dense layers and channels in convolutional layers, and facilitate \textit{group residual learning} by imposing partially ordered relation on these groups.
Specifically, if one group participates in the forward pass of model computation, then all of its preceding groups in this layer are also activated under such a structural constraint.
Therefore, we can use a single parameter \textit{slice rate} $r$ to control the proportion of groups involved in the forward pass during inference.
We empirically share the slice rate among all layers in the network; thus the computational resources required can be regulated precisely by the \textit{slice rate}.

\begin{figure}[h!]
    \centering
        \includegraphics[width=0.26\textwidth]{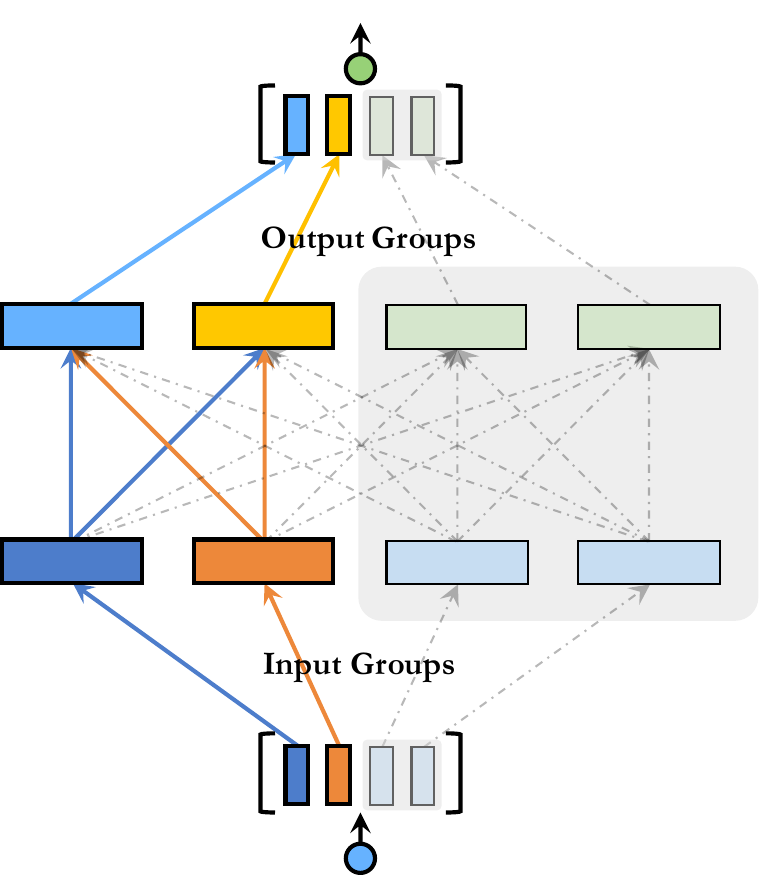}
    \caption{
    \textit{\TopicWord[M]}: dynamically slicing a sub-layer that is composed of preceding groups of the full layer controlled by the \textit{slice rate} $r$ (e.g., 0.5 here) during each forward pass.
    }
    \label{fig:multiplication}\vspace{-3mm}
\end{figure}

The \textit{slice rate} is structurally the same concept as \textit{width multiplier} \cite{howard2017mobilenets} which controls the width of the network.
However, instead of training only one fixed narrower model as in \cite{howard2017mobilenets}, we train the network in a dynamic manner to enhance the representation capacity of all the subnets it subsumes.
For each forward pass during training, as illustrated in Figure~\ref{fig:multiplication}, we sample the \textit{slice rate} from a distribution $F$ predetermined in the \textit{Slice Rate Scheduling Scheme}, and train the corresponding sub-layers.
The main challenges of training one model that supports inference at different widths include: how to determine proper candidate subnets (i.e. scheduling the \textit{slice rate}) for each training iteration; and more importantly, how to stabilize the scale of output for each component (i.e. neurons or channels) as the number of input components varies.
Independent to our work, \textit{Slimmable Neural Network}~\cite{yu2018slimmable} (\textit{SlimmableNet}) also proposes to train a single network executable at different widths.
In ~\cite{yu2018slimmable}, candidate subnets are considered to be equally important during training, by \textit{statically} scheduling \textit{all} subnets for every training pass and incorporating a set of batch normalization~\cite{ioffe2015batch} (BN) layers into each layer, one for each candidate sub-layer, to address the output scale instability issue.
In contrast, we consider the importance of the subnets to be different in \textit{\TopicWord[m][s]} (e.g., the full and the base network are the two most important subnets), and propose to dynamically schedule the training accordingly; besides the multi-BN solution, we further propose a more efficient solution with the group normalization\cite{wu2018group} layer (GN) to prevent the scale instability, which works in accordance with the dynamic group-wise training and engenders the \textit{group residual representation}.
We shall provide more discussions in Section~\ref{sec:network_slicing}.

The \textit{\TopicWord[m][s]} training scheme can be scrutinized under the perspective of residual learning~\cite{he2016deep,he2016identity} and knowledge distillation~\cite{hinton2015distilling}. 
Under the random training process of \textit{\TopicWord[m][s]}, \textit{groups} of each layer need to build up the representation increasingly, where the preceding groups carry the most fundamental information and the following groups the residual representation relatively.
Structurally, the final learned network is an ensemble of $G$ subnets, with $G$ being the number of groups, each corresponds to one \textit{slice rate}.
The parameters of these subnets are tied together and during each forward training pass, one subnet uniquely indexed by the slice rate is selected and trained.
We conjecture that the accuracy of the resulting fully trained network should be comparable to the network trained conventionally. 
Meanwhile, smaller subnets gradually distill knowledge from larger subnets as the training progresses, and thus can achieve comparable or even higher accuracy than their counterparts individually trained.
Consequently, we can provide the same functionality of an ensemble of models with only one model by width slicing.

The proposed training scheme has many advantages over existing methods on various issues such as model compression, model cascade and anytime prediction.
First, \textit{\TopicWord[m]} is readily applicable to existing neural networks, requiring no iterative retraining or dedicated library/hardware support as compared with most compression methods~\cite{han2015learning,liu2017learning}.
Second, instead of training a set of models and optimize the scheduling of these models with different accuracy-efficiency trade-offs as is in conventional model cascade~\cite{kang2017noscope,wang2017idk}, \textit{\TopicWord[m]} provides the same functionality of producing an approximate low-cost prediction with one single model.
Third, the structure of the model trained with \textit{\TopicWord[m]} naturally supports applications where the model is required to give prediction within a given computational budget dynamically, e.g., anytime prediction~\cite{huang2017multi,hu2019learning}.

Our main technical contributions are:

\begin{itemize}
    \item We develop a general training and inference framework \textit{\TopicWord[m][s]} that enables deep neural network models to support complex analytics with the trade-off between accuracy and inference cost/resource constraints on a per-input basis.
    
    \item We formally introduce the group residual learning of \textit{\TopicWord[m][s]} to general neural network models and further convolutional and recurrent neural networks. We also study the training details of \textit{\TopicWord} and their impact in depth.
    
    
    \item 
    We empirically validate through extensive experiments that neural networks trained with \textit{\TopicWord} can achieve performance comparable to an ensemble of networks with one single model and support fluctuating workload with up to 16x volatility.
    Example applications are also provided to illustrate the usability of \textit{\TopicWord}.
    The code is available at GitHub~\footnote{ \href{https://github.com/ooibc88/modelslicing}{https://github.com/ooibc88/modelslicing}}, which has been included in~\cite{ooi2015singa}.
\end{itemize}



The rest of the paper is organized as follows. Section~\ref{sec:related} provides a literature survey of related works.
Section~\ref{sec:network_slicing} introduces \textit{\TopicWord} and how it can be applied to various deep learning models, including Convolutional Neural Networks (CNNs), Recurrent Neural Networks (RNNs) and etc.
We then show how \textit{\TopicWord} can support fine-grained system degradation management for present industrial deep learning services and we also provide an illustrating application of cascade ranking in Section~\ref{sec:app}.
Experimental evaluations of \textit{\TopicWord} are given in Section~\ref{sec:experiment}, under prevailing natural language processing and computer vision tasks on public benchmark datasets.
Visualizations and detailed discussions of the results are also provided. 
Section~\ref{sec:conclusion} concludes the paper and points out some further research directions.

\section{Related Work}
\label{sec:related}


\subsection{Resource-aware Model Optimization}

Many recent works directly devise networks~\cite{huang2017multi,wang2018skipnet,cai2019isbnet} that are more economical in producing predictions.
SkipNet~\cite{wang2018skipnet} incorporates reinforcement learning into the network design, which guides the gating module on whether to bypass the current layer for each residual block.
SkipNet can provide predictions more efficiently yet in a less controlled manner inherently.
In MoE~\cite{shazeer2017outrageously}, a gating network is introduced to select a smaller number of networks out a mixture-of-experts which consists of up to thousands of networks during inference for each sample.
This kind of model ensemble approach aims to scale up the model capacity without introducing much overhead, while our approach enables every single model trained to scale down and support elastic inference cost.

MSDNet~\cite{huang2017multi} supports classification with computational resource budgets at test time by inserting multiple classifiers into a 2D multi-scale version of DenseNet~\cite{huang2017densely}. By early-exit into a classifier, MSDNet can provide predictions within given computation constraints.
ANNs~\cite{hu2019learning} adopts a similar design strategy of introducing auxiliary classifiers with Adaptive Loss Balancing, which supports the trade-off between accuracy and computational cost by using the intermediate features.
~\cite{mcintosh2018recurrent} also develops a model that can successively improve prediction quality with each iteration but this approach is specific to segmenting videos with RNN models.
These methods can largely alleviate the computational efficiency problem.
However, they are highly specialized networks, which restrict their applicability.
Functionally, models trained with \textit{\TopicWord[m]} also reuse intermediate features and support progressive prediction but with width slicing.
\textit{\TopicWord[M]} works similarly to these networks yet is more efficient, flexible and general.


\subsection{Model Compression}

Reducing the model size and computational cost has become a central problem in the deployment of deep learning solutions in real-world applications.
Many works have been proposed to resolve the challenges of growing network size and surging resource expenditure incurred, mainly memory and computation.
The mainstream solutions are to compress networks into smaller ones, including low-rank approximation~\cite{denton2014exploiting}, network quantization~\cite{courbariaux2016binarized,han2015deep,han2015learning}, weight pruning~\cite{han2015learning,han2015deep}, network sparsification on different level of structure~\cite{wen2016learning,liu2017learning} etc.

To this end, many model compression approaches attempt to reduce the model size on the trained networks.
~\cite{denton2014exploiting} reduces model redundancy with tensor decomposition on the weight matrix.
~\cite{courbariaux2016binarized} and~\cite{han2015learning} instead propose to quantize the network weights to save storage space. 
HashNet~\cite{chen2015compressing} also proposes to hash network weights into different groups and sharing weight values within each group. These techniques are effective in reducing model size. 
For instance,~\cite{han2015learning} achieves up to 35x to 49x compression rates on AlexNet~\cite{krizhevsky2012imagenet}. 
Although a considerable amount of storage can be saved, these techniques can hardly reduce run-time memory or inference time, and they typically need a dedicated library and/or hardware support.

Many studies propose to prune weights, filters or channels in the networks.
These approaches are generally effective because typically, deep networks are highly redundant in model representation.
~\cite{han2015deep,han2015learning} iteratively prune unimportant connections of small weights in trained neural networks.
~\cite{srinivas2017training} further guides the sparsification of neural networks during training by explicitly imposing sparse constraints over each weight with a gating variable. 
The resulting networks are highly sparse, which can be stored compactly in a sparse format.
However, the speedup of inference time of these methods depends heavily on dedicated sparse matrix operation libraries or hardware, and the saving of run-time memory is again very limited since most of the memory consumption comes from the activation maps instead of these weights.
~\cite{wen2016learning,liu2017learning} reduce the model size more radically by imposing regularization on the channel or filter and then prune the unimportant components.
Like \textit{\TopicWord[m][s]}, channel and filter level sparsity can reduce the model size, run-time memory footprint and also lower the number of computational operations.
However, these methods often require iterative fine-tuning to regain performance and support no inference time control.


\subsection{Efficient Model Design}

Instead of compressing existing large neural networks during or after training, 
recent works have also been exploring more efficient network design. 
ResNet~\cite{he2016deep,he2016identity} proposes residual learning via an identity mapping shortcut and the efficient bottleneck structure, which enables the training of very deep networks without introducing more parameters.
~\cite{veit2016residual} shows that ResNet behaves like an ensemble of shallow networks and it can still function normally with a certain fraction of layers being removed.
FractalNet~\cite{larsson2016fractalnet} contains a series of the duplication of the fractal architecture with interacting subpaths.
FractalNet adopts drop-path training which randomly selects certain paths during training, allowing for the extraction of fixed-depth subnetworks after training without significant performance loss.
To some extent, these network architectures can support on-demand workload by slicing subnets layer-wise or path-wise.
However, these methods are not generally applicable to other networks and the accuracy significantly drops when shortening or narrowing the network.

Many recent works focus on designing lightweight networks. 
SqueezeNet~\cite{iandola2016squeezenet} reduces parameters and computation with the fire module. 
MobileNet~\cite{howard2017mobilenets} and Xception~\cite{chollet2017xception} utilize depth-wise and point-wise convolution for more parameter efficient convolutional networks. 
ShuffleNet~\cite{zhang2018shufflenet} proposes point-wise group convolution with channel shuffle to help the information flowing across channels. 
These architectures scrutinize the bottleneck in conventional convolutional neural networks and search for more efficient transformation, reducing the model size and computation greatly.

\section{MODEL SLICING}
\label{sec:network_slicing}

We aim to provide a general training scheme for neural networks to support on-demand workloads with elastic inference costs.
More specifically, the target is to enable the neural network to produce prediction within a prescribed computational resource budget for each input instance, 
and meanwhile maintain the accuracy.

Existing methods of model compression, model ensemble and anytime prediction models can partially address this problem, but each has its limitations.
Model compression methods such as \textit{network slimming}~\cite{liu2017learning} which compresses channel width each layer, produce efficient models while they typically take longer training time for iterative pruning and retraining, and more importantly, have no control over resources required during inference.
Model ensemble methods, e.g., the ensemble of varying depth or width networks, support inference time resources control by scheduling the model for the immediate prediction task.
However, deploying an ensemble of the models multiply the amount of disk storage and memory consumption; further, scheduling of these models is a non-trivial task to the system in deployment.
Many works~\cite{huang2017multi,hu2019learning,mcintosh2018recurrent} instead exploit intermediate features for faster approximate prediction.
For instance, Multi-Scale DenseNet~\cite{huang2017multi} (MSDNet) inserts multiple classifiers into the model and thus supports anytime prediction by early-exit on a classifier.

\begin{figure}[t]
    \centering
        \includegraphics[width=0.48\textwidth]{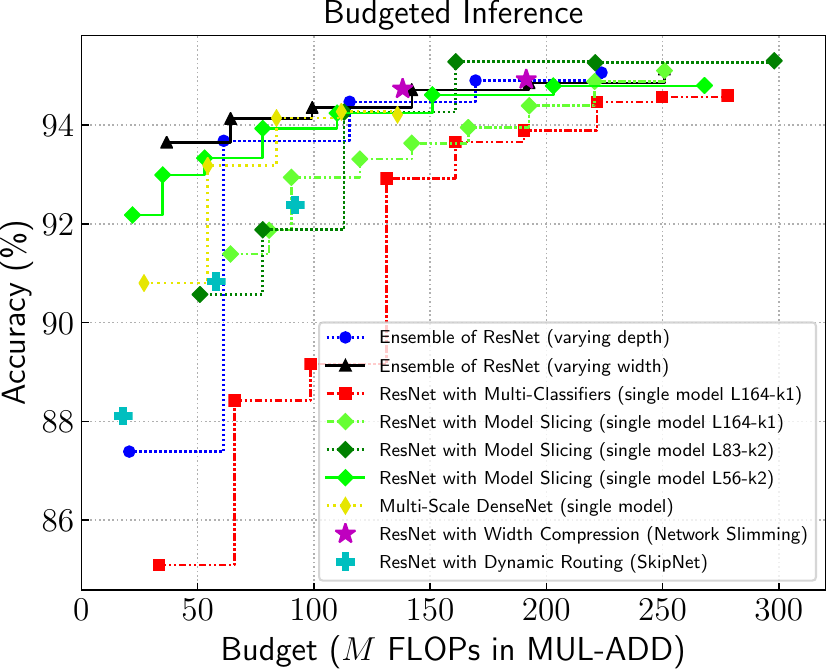}
    \caption{Classification accuracy w.r.t. inference FLOPs of ResNet trained with \textit{\TopicWord[m][s]} against ensemble, compression and other baselines on the CIFAR-10 dataset.
    }
    \label{fig:resnet_flops}\vspace{-8mm}
\end{figure}

Our \textit{\TopicWord[m][s]} also exploits and reuses intermediate features produced by the model while sidesteps the aforementioned problems.
The key idea is to develop a general training and inference mechanism called \textit{\TopicWord[m][s]} which slices a narrower subnet for faster computation.
With \textit{\TopicWord[m][s]}, neural networks are able to dynamically control the width of the subnet and thus regulate the computational resource consumption with one single parameter \textit{slice rate}.
In Figure~\ref{fig:resnet_flops}, we illustrate by comparing the accuracy-efficiency trade-offs of ResNet trained with different approaches.
We can observe that model ensemble methods are strong baselines that trade off accuracy for lower inference cost and that the Ensemble of ResNet with varying width performs better than varying depth.
This finding indicates the superiority of width slicing over depth slicing, which is corroborated by the rapid loss in accuracy of ResNet with Multi-Classifiers (single model) in Figure~\ref{fig:resnet_flops}.
We will show that trained with \textit{\TopicWord[m][s]}, one single model is able to provide inference performance comparable to the ensemble of varying width networks.
Therefore, \textit{\TopicWord[m][s]} is an ideal solution for neural networks to support elastic inference cost and resource constraints.

\subsection{Model Slicing for Neural Networks}
\label{sec:model_slicing_nn}

We start by introducing \textit{\TopicWord[m][s]} to fully-connected layer (dense layer) for general neural networks.
Each dense layer in the neural network transforms via a weight matrix $\mathbf{W} \in \mathbb{R}^{N \times M}$: $\mathbf{y} = \mathbf{W} \mathbf{x}$, where $\mathbf{x} = [x_1, x_2, \ldots, x_M]$, a $M$-dimension input vector, corresponds to $M$ input neurons and $\mathbf{y} = [y_1, y_2, \ldots, y_N]$, $N$ output neurons correspondingly. Details such as the bias and non-linearity are omitted here for brevity. 
As illustrated in Figure~\ref{fig:multiplication}, a gating variable is \textit{implicitly} introduced to impose a structural constraint on each input neuron $x_j$:

\begin{equation}
\label{formular:nn_op}
    y_i = \sum_{j=1}^{M} w_{ij} (\alpha_j \cdot x_j)
\end{equation}

Each gating variable $\alpha_j$ thus controls the participation of the corresponding neuron $x_j$ in each forward pass during both training and inference.
Formally, the structural constraint is obtained by imposing partial ordered relation on these gating variables:

\begin{equation}
\label{formular:constraint}
    \forall i \forall j (i < j \wedge \alpha_j=1 \rightarrow \alpha_i=1)
\end{equation}

which requires that the set of activated neurons during each forward pass forms a contiguous block starting from the \textit{first} neuron.
Based on the relation, we further divide these neurons into $G$ ordered groups, i.e. $\mathbf{x} = [\mathbf{x_1}, \mathbf{x_2}, \ldots, \mathbf{x_G}]$, each group corresponds to a contiguous block of neurons.
We denote the index of the rightmost neuron of the \textit{first $i$ groups} as $g_i$, and the corresponding sub-layer as Sub-layer-$r_i$, where the \textit{slice rate} $r_i = \frac{g_i}{M}, (0 < r_i \le 1)$.
Then the set of groups involved in the current forward pass can be determined by indexing the rightmost group $\mathbf{x_i}$, and the set of neurons involved corresponds to $\{x_1, x_2, \ldots, x_{g_i}\}$.
Note that the group number $G$ is a pre-defined hyper-parameter, which could be set from 1 (the original layer) to $M$ (each component forms a group).

Empirically, the \textit{slice rate} is shared among all the layers in the network and we denote the subnet of first $i$ groups in each layer as Subnet-$r_i$.
Thus the width of the whole network can be regulated by the single parameter $r$.
As illustrated in Figure~\ref{fig:multiplication}, only the sliced part of the weight matrix and components are activated and required to reside in memory for inference in the current forward pass.
We denote the computational operation required by the full network as $C_0$, then the computational operation required by the subnet of \textit{slice rate} $r$ is roughly $r^2 \times C_0$.
Therefore, the run-time computational resources limit $C_t$ can be dynamically satisfied by restricting \textit{slice rate} $r$ by:

\begin{equation}
\label{formular:slice_rate}
    r \leq min(\sqrt{\frac{C_t}{C_0}}, 1)
\end{equation}

Consequently, a subnet can be readily sliced and deployed out of the network trained with \textit{\TopicWord[m][s]} whose disk storage and run-time memory consumption are also roughly quadratic to the \textit{slice rate} $r$.
Besides satisfying the run-time computational constraint, another primary concern is how to maintain the performance of these subnets.
To this end, we propose the \textit{\TopicWord[m][s]} training in Algorithm~\ref{algo:model_slicing}.
For each training pass, a list of slice rate $\mathbf{L}_t$ is sampled from the predefined slice rate list $\mathbf{L}$ by a scheduling scheme $\mathcal{F}$, and the corresponding subnets are optimized under the current training batch.
We shall elaborate on the scheduling scheme in Section~\ref{sec:scheduling_scheme}.

\begin{algorithm}[t]
    \caption{Training with Model Slicing.}
    \label{algo:model_slicing}
    \textbf{Input:} model $\mathbf{W}_0$, slice rate list $\mathbf{L}$, scheduling scheme $\mathcal{F}$, training iteration $T$, \textit{criterion}, \textit{optimizer}.\\
    
    Upgrade layers
    to support \textit{model slicing}: $\mathbf{W}_0 \leftarrow upgrade\_model(\mathbf{W}_0, \mathbf{L})$ \\
    
    \For{iteration $t$ \KwFrom $0$ \KwTo $T-1$}{
        Generate next batch of data and label: $(\mathbf{x}_t, \mathbf{y}_t)$   \\
        Generate the current training slice rate list: 
            $\mathbf{L}_t \leftarrow next\_slice\_rate\_batch(\mathbf{L}, \mathcal{F})$   \\
        Initialize model gradient $\mathbf{W}_g \leftarrow 0$    \\
        \For{slice rate  $r \in \mathbf{L}_t$}{
            Forward Subnet-$r$: $\hat{\mathbf{y}} \leftarrow forward(\mathbf{W}_t, r, \mathbf{x}_t)$   \\
            Compute Loss: $loss \leftarrow criterion(\mathbf{y}_t, \hat{\mathbf{y}})$  \\
            Accumulate gradient: $\mathbf{W}_g \leftarrow \mathbf{W}_g + loss.backward()$
        }
        Update model $\mathbf{W}_{t+1} \leftarrow optimizer.update(\mathbf{W}_t, \mathbf{W}_g)$\\
    }
\end{algorithm}

Notice that the parameters of all subnets are tied together and any subnet indexed by a slice rate $r_i$ subsumes all smaller subnets.
The structural constraint of \TopicWord[m][s] is reminiscent of residual learning~\cite{he2016deep,he2016identity}, where the Subnet-$r_1$ (the base network) carries the base representation.
With the new input group $\mathbf{x_i}$ introduced as $i$ grows, each $y_j$ is optimized to learn from finer input details and thus the \textit{group residual presentation}.
We shall provide more discussions on this effect in Section~\ref{sec:residual_learning_effect}.
From the viewpoint of knowledge distillation~\cite{hinton2015distilling}, the Subnet-$r_G$ (Subnet-$1.0$) maintains the capacity of the full model and as the training progresses, each Subnet-$r_i$ gradually distills the representation from larger subnets and transfers the knowledge to smaller ones.
Under this training scheme, we conjecture that the full network can maintain the accuracy, or possibly improve due to the regularization and ensemble effect; and in the meantime, the subnets can gradually pick up the performance by distilling knowledge from larger subnets.

\subsection{Convolutional Neural Networks}
\label{sec:model_slicing_cnn}

\textit{\TopicWord[M][s]} is readily applicable to convolutional neural networks in a similar manner.
The most fundamental operation in CNNs comes from the convolutional layer which can be constructed to represent any given transformation $\mathcal{F}_{conv}:\mathbf{X} \rightarrow \mathbf{Y}$, where $\mathbf{X} \in \mathbb{R}^{M \times W_{in}\times H_{in}}$ is the input with $M$ channels of size $W_{in}\times H_{in}$, $\mathbf{Y} \in \mathbb{R}^{N \times W_{out}\times H_{out}}$ the output likewise. 
Denoting $\mathbf{X} = [\mathbf{x}_1, \mathbf{x}_2, \ldots, \mathbf{x}_{M}]$ and $\mathbf{Y} = [\mathbf{y}_1, \mathbf{y}_2, \ldots, \mathbf{y}_{N}]$ in vector of channels, the parameter set associated with each convolutional layer is a set of filter kernels $\mathbf{K} = [\mathbf{k}_1, \mathbf{k}_2, \ldots, \mathbf{k}_{N}]$. 
In a way similar to the dense layer, \textit{\TopicWord[m][s]} for the convolutional layer can be represented as:

\begin{equation}
\label{formular:cnn_op}
    \mathbf{y}_i = \mathbf{k}_i \ast \mathbf{X} = \sum_{j=1}^{M} \mathbf{k}_{i}^{j} \ast (\alpha_j \cdot \mathbf{x}_j)
\end{equation}

where $\ast$ denotes convolution operation, $\mathbf{k}_i^j$ is a 2D spatial kernel associated with $i_{th}$ output channel $\mathbf{y}_i$ and convolves on $j_{th}$ input channel $\mathbf{x}_j$.
Consequently, treating channels in convolutional layers analogously to neurons in dense layers, \textit{\TopicWord[m][s]} can be directly applied to CNNs with the same training scheme.

Nonetheless, the output scale instability issue arises when applying \textit{\TopicWord[m][s]} to CNNs.
Specifically, each convolutional layer is typically coupled with a batch normalization layer~\cite{ioffe2015batch} to normalize outputs in the batch dimension, which stabilizes the mean and variance of input channels received by channels in the next layer.
In the implementation of Equation~\ref{formular:batch_norm}, each batch-norm layer normalizes outputs with the batch mean $\mu$ and variance $\sigma^2$ and keeps records of running estimates of them which will be used directly after training. 
Here, $\gamma$ and $\beta$ are learnable affine transformation parameters of this batch-norm layer associated with each channel.
However, with \TopicWord[m][s], the number of inputs received by a given output channel is no longer fixed, which is instead determined by the slice rate $r_i$ during each forward pass.
Consequently, the mean and variance of the batch-norm layer on the output fluctuate drastically; thus \textit{one single set} of the running estimates is unable to stabilize the distribution of the output channel.

\begin{equation}
\label{formular:batch_norm}
    \mathbf{\hat{y}} = \frac{\mathbf{y}_{in}-\mu}{\sqrt{\sigma^2+\epsilon }}; \mathbf{y}_{out}=\gamma \mathbf{\hat{y}} + \beta 
\end{equation}

We propose to address this issue with Group Normalization~\cite{wu2018group}, an adaptation to Batch-norm.
Group-norm divides channels into groups and normalizes channels in the same way as is in Equation~\ref{formular:batch_norm} with the only difference that the mean and variance are calculated dynamically within each group.
Formally, given the total number of groups $G$ , the mean $\mu_i$ and variance $\sigma_i^2$ of $i$-th group are estimated within the set of channels in Equation~\ref{formular:group_norm} 
and shared among all the channels in the $i$-th group for normalization.

\begin{equation}
\label{formular:group_norm}
    \mathcal{S}_i = \{ \mathbf{x}_j | floor(\frac{j-1}{G}) = i \}
\end{equation}

Group-norm normalizes channels group-wise instead of batch-wise, avoiding running estimates of the batch mean and variance in batch-norm whose error increases rapidly as the batch size decreases.
Experiments in~\cite{wu2018group}, which is also validated by our experiments on various network architectures, show that the accuracy of group-norm is relatively stable with respect to the batch size and group number.
Besides stabling the scale, another benefit of group-norm is that it engenders the group-wise representation, which is in line with the \textit{group residual learning} effect of \TopicWord[m][s] training.
To introduce \TopicWord[m][s] to CNNs, we only need to replace batch-norm with group-norm and slice the normalization layers together with convolutional layers at the granularity of the group.

\subsection{Recurrent Neural Networks}
\label{sec:model_slicing_rnn}

\textit{\TopicWord[M][s]} can be readily applied to recurrent layers similarly to fully-connected layers.
Take the vanilla recurrent layer expressed in Equation~\ref{formular:rnn_op} for demonstration, the difference is that the output $\mathbf{h}_t$ is computed from two sets of inputs, namely $\mathbf{x}_t$ and $\mathbf{h}_{t-1}$.

\begin{equation}
\label{formular:rnn_op}
     \mathbf{h}_t = \sigma (\mathbf{W}_{hx}\mathbf{x}_t+\mathbf{W}_{hh}\mathbf{h}_{t-1}+\mathbf{b}_h)
\end{equation}

Consequently, we can slice each input of the recurrent layer separately and adopt the same training scheme as fully-connected layers.
\TopicWord[M][s] for recurrent layers of RNN variants such as GRU~\cite{cho2014properties} and LSTM\cite{hochreiter1997long} works similarly.
Dynamic slicing is applied to all input and output sets, including hidden/memory states and various gates, regulated by one single parameter \textit{slice rate} $r$ of each layer.

\subsection{Slice Rate Scheduling Scheme}
\label{sec:scheduling_scheme}

As shown in Algorithm~\ref{algo:model_slicing}, for each training pass of \TopicWord[m][s], a list of \textit{slice rate} is sampled from a predetermined scheduling scheme $\mathcal{F}$, and then the corresponding subnets are trained under the current training batch.
Formally, the random scheduling can be described as sampling the \textit{slice rate} $r$ from a Distribution $\mathcal{F}$.
Denoting the list of valid \textit{slice rate} $r$ in order as $(r_1, r_2, \dots, r_G)$, then we have:

\begin{equation}
\resizebox{.9\hsize}{!}{$
  \begin{cases}
    p(r_1) = F( \frac{r_1+r_2}{2} ) =  \int_{- \infty}^{ \frac{r_1+r_2}{2}} f(r)dr, & i=1\\
    p(r_i) = F( \frac{r_i+r_{i+1}}{2} ) - F( \frac{r_{i-1}+r_{i}}{2} ) =  \int_{\frac{r_{i-1}+r_{i}}{2}}^{ \frac{r_i+r_{i+1}}{2}} f(r)dr, & 1 < i < G\\
    p(r_G) = 1-F( \frac{r_{G-1}+r_{G}}{2} ) =  \int_{\frac{r_{G-1}+r_{G}}{2}}^{ +\infty } f(r)dr, & i=G
  \end{cases}
$}
\end{equation}

where $f(r)$ is the probability density function, $F(r)$ the cumulative distribution function of $\mathcal{F}$ and $p(r_i)$ the probability of \textit{slice rate} $r_i$ being sampled.
Thereby, the random scheduling $\mathcal{F}$ (e.g., the Uniform Distribution or the Normal Distribution) can be parameterized with a Categorical Distribution $Cat(G, p(r_1), p(r_2), \dots, p(r_G))$, where each $p(r_i)$ denotes the relative importance of Subnet-$r_i$ over other subnets.
Further, the importance of these subnets should be treated differently.
In particular, the full and the base network (i.e. Subnet-$r_G$ and Subnet-$r_1$) should be the two most important subnets, because the full network represents the model capacity and the base network forms the basis for all the subnets.
Based on this observation, we propose three categories of scheduling schemes:

\begin{itemize}
	\item \textit{Random scheduling}, where \textit{each} of the slice rate is sampled from an $\mathcal{F}$ parameterized by $(p(r_1), \dots, p(r_G))$.
	
	\item \textit{Static scheduling}, where \textit{all} valid slice rates are scheduled for the current training pass.
	
	\item \textit{Random static scheduling}, where both a \textit{fixed} set and a set of randomly sampled slice rates are scheduled.
\end{itemize}

For \textit{random scheduling}, the importance of different subnets can be represented in the assigned probabilities, where we can assign higher sampling probabilities to more important subnets (e.g., the full and base network) during training.
Likewise, for \textit{random static scheduling}, we can include the important subnets in the fixed set and meanwhile assign proper probabilities to the remaining subnets.
We shall evaluate these \textit{slice rate} scheduling schemes in Section~\ref{sec:scheduling_exp}.

\subsection{Group Residual Learning of Model Slicing}
\label{sec:residual_learning_effect}

The \textit{\TopicWord} training scheme structurally is reminiscent of residual learning proposed in ResNet~\cite{he2016deep,he2016identity}.
In ResNet, a shortcut connection of identity mapping is proposed to forward input to output directly: $\mathbf{y} = \mathbf{x} + \mathcal{F}_{conv}(\mathbf{x})$, where during optimization, the convolutional transformation only needs to learn the residual representation on top of input information $\mathbf{x}$, namely $\mathbf{y}-\mathbf{x}$.
Analogously, networks trained with \TopicWord[m][s] learn to accumulate the representation with additional groups introduced (group of neurons in dense layers and group of channels in convolutional layers), i.e. $\mathbf{y} = \sum_{i=1}^G \mathcal{F}_{conv\_i}(\mathbf{x_i})$.

To demonstrate the group residual learning effect in \textit{\TopicWord}, we take the transformation in a fully-connected layer for example, and analyze the relationship between any two sub-layers of \textit{slice rate} $r_a$ and $r_b$ with $r_a < r_b$.
We have the transformation of Sub-layer-$r_a$ as $\mathbf{y}_a = \mathbf{W}_a \mathbf{x}_a$ and the transformation of Sub-layer-$r_b$ $[\mathbf{\tilde{y}}_a; \mathbf{y}_b] = \mathbf{W}_b [\mathbf{x}_a; \mathbf{x}_b]$ in block matrix multiplication as:

\begin{gather}
    \begin{bmatrix} \mathbf{\tilde{y}}_a \\ \mathbf{y}_b \end{bmatrix} =
    \begin{bmatrix}
    \mathbf{W}_a & \mathbf{B} \\
    \mathbf{C} & \mathbf{D}
    \end{bmatrix} \cdot
    \begin{bmatrix}
    \mathbf{x}_a \\ \mathbf{x}_b
    \end{bmatrix} =
    \begin{bmatrix}
    \mathbf{W}_a \mathbf{x}_a + \mathbf{B} \mathbf{x}_b \\
    \mathbf{C} \mathbf{x}_a + \mathbf{D} \mathbf{x}_b
    \end{bmatrix}
\end{gather}

Here, $\mathbf{x}_b$ is the supplementary input group introduced for Sub-layer-$r_b$ and $\mathbf{y}_b$ is the corresponding output group.
Generally $r_b-r_a \ll r_a$, then the \textit{group residual representation learning} can be clarified from two angles.
Firstly, the base representation of Sub-layer-$r_b$ is $\mathbf{\tilde{y}}_a = \mathbf{W}_1 \mathbf{x}_a + \mathbf{B} \mathbf{x}_b = \mathbf{y}_a + \mathbf{B} \mathbf{x}_b$, which is composed of the base representation $\mathbf{y}_a$ and the residual representation $\mathbf{B} \mathbf{x}_b$.
Secondly, the newly-introduced output group $\mathbf{y}_b$ further forms the residual representation supplementary to the base representation $\mathbf{\tilde{y}}_a$.
Higher model capacity is therefore expected of Subnet-$r_b$.

The justification for the \textit{group residual learning effect} in \textit{\TopicWord} is that as the training progresses, the base representation of $\mathbf{y}_a$ alone in Sub-layer-$r_a$ has already been optimized for the learning task.
Therefore, the supplementary group $\mathbf{y}_b$ introduced to Sub-layer-$r_b$ gradually adapts to learn the residual representation, which is corroborated in the visualization in Section~\ref{sec:residual_learning_vis}.
Furthermore, this group residual learning characteristic provides an efficient way to harness the richer representation for Subnet-$r_b$ based on Subnet-$r_a$ by the simple approximation of $\mathbf{\tilde{y}}_a \approx  \mathbf{y}_a$.
With this approximation in every layer of the network, the most computationally heavy features of $\mathbf{W}_a \mathbf{x}_a$ could be reused without re-evaluating, thus the representation of Sub-layer-$r_b$ can be updated by calculating only $\mathbf{C} \mathbf{x}_a + \mathbf{D} \mathbf{x}_b$ with a significantly lower computational cost.

We note that the \textit{\TopicWord[m][s]} training for \textit{group residual representation} is applicable to the majority of neural networks.
In addition, the \textit{group residual learning} mechanism of \textit{\TopicWord[m][s]} is ideally suited for networks with layer transformation of multiple branches, e.g., group convolution~\cite{zhang2018shufflenet}, depth-wise convolution~\cite{howard2017mobilenets} and homogeneous multi-branch residual transformation of ResNeXt~\cite{xie2017aggregated} etc.

\section{Example Applications}
\label{sec:app}

In this section, we demonstrate how \textit{\TopicWord[m][s]} can benefit the deployment of deep learning based services. 
We use \textit{\TopicWord[m][s]} as the base framework to manage fine-grained system degradation for large scale machine learning services of dynamic workload.
We also provide an example application of cascade ranking with \textit{\TopicWord[m][s]}.

\subsection{Supporting Dynamic Workload Services}
\label{sec:app_sys_degradation}

For a service with a dynamic workload, fine-grained system degradation management can be supported directly and efficiently with \textit{\TopicWord[m][s]}.
Query samples come as a stream, and there is a dynamic latency constraint.
Queries are usually batch-processed with vectorized computation for higher efficiency.

We design and implement an example solution to guarantee the latency and throughput requirement via \textit{\TopicWord[m][s]}. 
Given the processing time per sample for the full model $t$, to satisfy the dynamic latency constraint $T$ and unknown query workload, we can build a mini-batch in every $T/2$ time, and utilize the rest $T/2$ time budget for processing: first examine the number of samples $n$ in current batch, and choose the slice rate $r$ satisfying $nr^2t \leq T/2$ (Equation~\ref{formular:slice_rate}) so that the processing time for this batch is within the budget $T/2$.
Under such a system design, no computation resource is wasted as the total processing time per mini-batch is exactly the time interval of the batch input.
Meanwhile, all samples can be processed within the required latency.

\subsection{\TopicWord[M][S] for Cascade Ranking Applications}
\label{sec:cascade_ranking}

Many information retrieval and data mining applications such as search and recommendation need to rank a large set of data items with respect to many user requests in an online manner. 
There are generally two issues in this process: 
1). Effectiveness as how accurate the obtained results in the final ranked list are and whether there are a sufficient number of good results; 
and 2). Efficiency such as whether the results are obtained in a timely manner from the user perspective and whether the computational costs of ranking is low from the system perspective. 
For large-scale ranking applications, it is of vital importance to address both issues for providing good user experience and achieving a cost-saving solution.

Cascade ranking~\cite{wang2011cascade,liu2017cascade} is a strategy designed for such a trade-off.
It utilizes a sequence of prediction functions of different costs in different stages.
It can thus eliminate irrelevant items (e.g., for a query) in earlier stages with simple features and models, while segregate more relevant items in later stages with more complicated features and models.
In general, functions in early stages require low inference cost while functions in later stages require high accuracy.

One critical characteristic of cascade ranking is that the optimization target for each function may depend on all other functions in different stages~\cite{liu2017cascade}.
For instance, given a positive item set $\{1,2,...,7\}$ and we aim to build a cascade ranking solution with two stages,
suppose that function in stage two mis-drop positive item $\{6,7\}$, a function in stage one mis-drop $\{1,6,7\}$ is better than a function mis-drop $\{1,2\}$, though the former has a higher error rate over the whole dataset (in the first case $\{2,3,4,5\}$ are left while in the second case only $\{3,4,5\}$ are left). Lots of analysis are given in~\cite{wang2011cascade,chen2017efficient,liu2017cascade}. 
Therefore, we expect the prediction of positive items given by functions in different stages to be consistent so that the accumulated false negatives are minimized.
Unfortunately, most implementations of the ranking/filtering function at each stage for cascade ranking use different model architectures with different parameters.
The results of different models are thus unlikely to be consistent.

\textit{\TopicWord[M][s]} would be an ideal solution for cascade ranking.
Firstly, it provides the trade-off of model effectiveness and model efficiency with one single model.
The ranking functions at different stages can be obtained by as simple as configuring the inference cost of the model.
Secondly, as is corroborated in Section~\ref{sec:visualization}, the prediction results of \textit{\TopicWord[m][s]} sub-models are inherently correlated since the larger model is actually using the smaller model as the base of its model representation.
We shall illustrate the effectiveness and efficiency of \textit{\TopicWord} in comparison with the traditional model cascade solution in a cascade ranking simulation in Section~\ref{sec:simulation}.

\section{Experiments}
\label{sec:experiment}

We evaluate the performance of \textit{\TopicWord} on state-of-the-art neural networks on two categories of public benchmark tasks, specifically evaluating \textit{\TopicWord[m][s]} for dense layers, i.e. fully-connected and recurrent layers on language modeling~\cite{mikolov2010recurrent,zaremba2014recurrent,press2016using} in Section~\ref{sec:exp_nnlm} and evaluating \textit{\TopicWord[m][s]} for convolutional layers on image classification~\cite{simonyan2014very,he2016deep,zagoruyko2016wide} in Section~\ref{sec:exp_cnn}.
Experimental setups of \textit{\TopicWord} are provided in Section~\ref{sec:experiment_setup}; cascade ranking simulation of example applications and visualization on the \textit{\TopicWord} training  
are given in Section~\ref{sec:simulation} and Section~\ref{sec:visualization} respectively.

\subsection{Model Slicing Setup}
\label{sec:experiment_setup}

\subsubsection{General Setup and Baselines}
\label{sec:setup}

The slice rate $r_i$ corresponds to Subnet-$r_i$, which is restricted between a lower bound $r_1$ and $1.0$.
In the experiments, the networks trained with \textit{\TopicWord} are evaluated with the slice rate list where $r_i$ ranges from $r_1=0.25/0.375$ (corresponding to around 16x/7x the computational speedup) to $1.0$ in every $\frac{1}{4}/\frac{1}{8}/\frac{1}{16}$ (the slice granularity).
We apply \textit{\TopicWord[m][s]} to all the hidden layers except the input and output layers because both layers are necessary for the inference and further take a negligible amount of parameter and computation in the full network.

We compare \textit{\TopicWord} primarily with two baselines.
The first baseline is the full network trained without \textit{\TopicWord} ($r_1=1.0$, \textit{single model}), implemented by fixing $r_1$ to $1.0$ during training.
During inference, we slice the corresponding Sub-layer-$r_i$ of each layer in the network for comparison.
The second baseline is an ensemble of networks of varying width (\textit{fixed models}).
In addition to the above two baselines, we also compare \TopicWord[m][s] with model compression (Network Slimming~\cite{liu2017learning}), anytime prediction (multi-classifiers methods, e.g. MSDNet~\cite{huang2017multi}) and efficient prediction (SkipNet~\cite{wang2018skipnet}).

\subsubsection{Slice Rate Scheduling Scheme}
\label{sec:scheduling_exp}

\begin{table}[h!]
    \centering
    \renewcommand{\arraystretch}{1.}
    \caption{Accuracy of VGG-13 trained with various training scheduling schemes on CIFAR-10. 
    $|\mathcal{L}_t|$ denotes the number of slice rates scheduled for each training pass.
    }
    \resizebox{\columnwidth}{!}{
        \begin{tabular}{ c || c|ccccccc|c}
    \thickhline
    
    Scheme & Fixed & R-uniform-2 & R-weighted-2  & R-weighted-3  & Static  & R-min  & R-max  & R-min-max & Slimmable \\
    $|\mathcal{L}_t|$  & 4 & 2 & 2 & 3 & 4 & 2 & 2 & 3 & 4 \\
    \hline
    1.00  & 94.31 & 93.72  & 94.23 & 94.34 & 93.67 & 93.15 & 94.32 & \textbf{94.35} & \textbf{\underline{94.41}} \\
    0.75  & 93.86 & 93.64  & 94.08 & \textbf{94.20} & 93.46 & 93.14 & 93.59 & 93.97 & \textbf{\underline{94.29}} \\
    0.50  & 93.39 & 93.68  & 93.76 & \textbf{\underline{93.92}} & 93.19 & 93.11 & 93.05 & 93.60 & 93.47 \\
    0.25  & 91.63 & 91.59 & 91.68 & 91.96 & 91.69 & 91.84 & 91.31 & \textbf{\underline{92.10}} & 91.45 \\
    
    \thickhline
        \end{tabular}
    }
    \label{tab:dist_comp}
\end{table}

We evaluate the three slice rate scheduling schemes proposed in Section~\ref{sec:scheduling_scheme} with the slice rate list $(1.0, 0.75, 0.5, 0.25)$ in Table~\ref{tab:dist_comp}.
Specifically, the baseline is the ensemble of fixed models (\textit{fixed}).
For \textit{random scheduling}, we evaluate the uniform sampling (\textit{R-uniform}) and the weighted random sampling (\textit{R-weighted}, weight list $(0.5, 0.125, 0.125, 0.25)$); in particular, \textit{R-uniform-k} and \textit{R-weighted-k} denote random scheduling of k slice rates scheduled for each forward pass.
For \textit{static scheduling} (\textit{Static}), the subnets are regarded as equally important and thus all slice rates are scheduled whose \textit{computation} grows linearly with the number of subnets configured;
For \textit{random static scheduling}, we evaluate statically scheduling the base network (\textit{R-min}), the full network (\textit{R-max}) or both of these two subnet (\textit{R-min-max}), and meanwhile uniformly sampling one remaining subnets.
The detailed training settings are given in Section~\ref{sec:cnn_training_details}.

Table~\ref{tab:dist_comp} shows that weighted sampling of \textit{random scheduling} achieves higher accuracy than uniformly sampling with a comparable training budget; and training longer further improves the performance.
In contrast, \textit{static scheduling} performs consistently worse than the weighted \textit{random scheduling} even though it takes more training rounds.
The results corroborate our conjuncture that the base and the full network are of greater importance and thus should be scheduled more frequently during training.

We next evaluate the \textit{random static scheduling}, which consists of statically scheduling the base and/or full network while uniformly sampling the remaining subnets.
We observe that statically training the base (\textit{R-min}) or the full (\textit{R-max}) network helps to improve the corresponding subnets.
Meanwhile, the performance of the neighboring subnets also improves, mainly due to the effect of knowledge distillation.
We also compare \textit{\TopicWord} with
\textit{SlimmableNet}~\cite{yu2018slimmable} (\textit{Slimmable}) that adopts \textit{static scheduling} and multi-BN layers instead of one group-norm layer.
The results shown in Table~\ref{tab:dist_comp} reveal that \textit{SlimmableNet} obtains higher accuracies in larger subnets, which may result from the longer training time; while smaller subnets perform worse than \textit{\TopicWord} with \textit{random scheduling}, e.g., \textit{R-weighted} or \textit{R-min-max}, mainly due to the lack of differentiation of varying importance of subnets in \textit{static scheduling}.
In the following experiments, we therefore evaluate \textit{\TopicWord} with \textit{R-weighted-3} for small datasets and \textit{R-min-max} for larger datasets for reporting purpose.

\subsubsection{The Lower Bound of Slice Rate}

\begin{figure}[h!]
    \centering
        \includegraphics[width=0.35\textwidth]{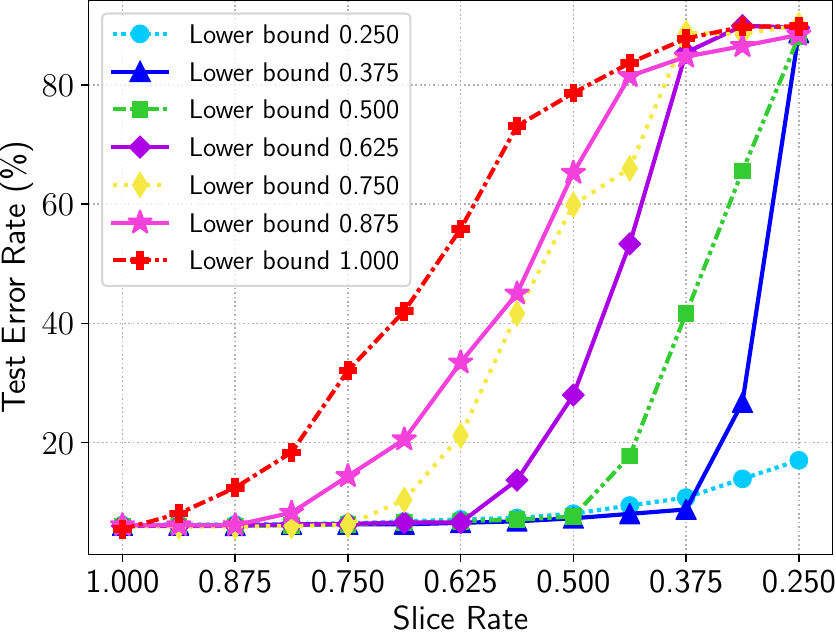}
    \caption{Illustration of the impact of the lower bound ($lb$) on VGG-13 trained with \textit{\TopicWord} on CIFAR-10.}
    \label{fig:lb_comp}
\end{figure}

For each of the subnet, the computation resources required can be evaluated beforehand.
The lower bound controls the width of the base network and thus should be set to Equation~\ref{formular:slice_rate} under the computational resource limit.
Figure~\ref{fig:lb_comp} shows the accuracies of VGG-13 trained with different lower bounds.
Empirically, the accuracy drops steadily as $r_i$ decreases towards $r_1$ (the lower bound $lb$), and networks trained with different $lb$s perform rather close.
Given a lower bound $lb$, however, the accuracy of the corresponding Subnet-$lb$ is slightly higher than other Subnet-$lb$s, which is mainly because the base network is optimized more frequently.
When the slice rate $r_i$ decreases over the lower bound, the accuracy drops drastically.
This phenomenon meets the expectation that further slicing the base network destroys the base representation, and thus the accuracy suffers significantly.
The loss of accuracy is more severe for convolutional neural networks, where the representation depends heavily on all channels of the base network.
In the following experiments, we therefore evaluate lower bound $0.375/0.25$ for small (e.g. CIFAR, PTB)/large (e.g. ImageNet) datasets respectively for reporting purpose, whose computational cost is roughly 14.1\%/6.25\% of the full network (i.e. 7.11x/16x speedup) and empirically can be adjusted readily according to the deployment requirement.



\subsection{NNLM for Language Modeling}
\label{sec:exp_nnlm}

\subsubsection{Language modeling task and dataset}
\label{sec:nnlm_setup}

The task of language modeling is to model the probability distribution over a sequence of words. 
Neural Network Language Modeling (NNLM) comprises both fully-connected and recurrent layers; we thus adopt NNLM to evaluate the effectiveness of \textit{\TopicWord[m][s]} for dense layers.
NNLM~\cite{mikolov2010recurrent,zaremba2014recurrent,press2016using} specifies the distribution over next word $w_{t+1}$ given its preceding word sequence $w_{1:t}=[w_1, w_2, \dots, w_t]$ with neural networks.
Training of NNLM involves minimizing the negative log-likelihood ($NLL$) of the sequence: $NLL = -\sum_{t=1}^{T}logP(w_t|w_{1:t-1})$.
Following the common practice for language modeling, we use perplexity ($PPL$) to report the performance: $PPL = exp(\frac{NLL}{T})$.
We adopt the widely benchmarked English Penn Tree Bank (PTB) dataset and use the standard train/test/validation split by~\cite{mikolov2010recurrent}.

\begin{figure}[h!]
    \centering
        \includegraphics[width=0.38\textwidth]{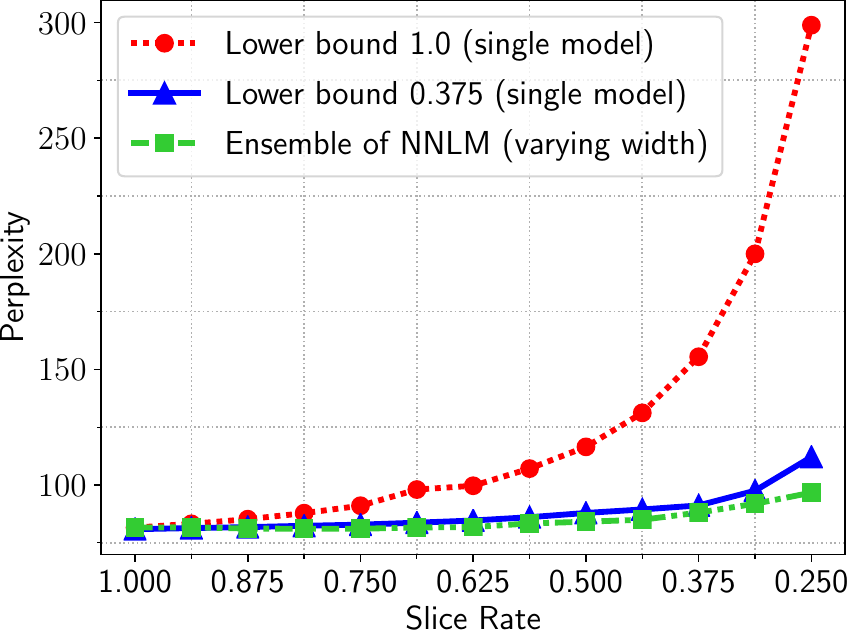}
    \caption{Results of NNLM trained w/o \textit{\TopicWord}.}
    \label{fig:lm_ppl}
\end{figure}

\subsubsection{NNLM configuration and training details}
Following~\cite{mikolov2010recurrent,zaremba2014recurrent,press2016using}, the NNLM model in the experiments consists of an input embedding layer, two consecutive LSTM layers, an output dense layer and finally a softmax layer.
The embedding dimension is $650$ and both LSTM layers contain $640$ units.
In addition, a dropout layer with a dropout rate $0.5$ follows the embedding and two LSTM layers.
The models are trained by truncated backpropagation through time for $35$ time steps, minimizing $NLL$ during training without any regularization terms with SGD of batch size $20$.
The learning rate is initially set to $20$ and quartered in the next epoch if the perplexity does not decrease on the validation set.
\textit{\TopicWord[M][s]} applies to both recurrent layers and the output dense layer with output rescaling.

\subsubsection{Results of Model Slicing on NNLM}

Results in Figure~\ref{fig:lm_ppl} and Table~\ref{tab:nnlm_ppl} show that \textit{\TopicWord} is effective to support on-demand workload with one single model only at the cost of minimum performance loss.
The performance of the network trained without \textit{\TopicWord} decreases drastically.
With \textit{\TopicWord}, the performance decreases steadily and stays comparable to the corresponding fixed models.
In particular, the performance of the subnet is slightly better than the corresponding fixed model when the \textit{slice rate} is near $1.0$.
For instance, as is shown in Table~\ref{tab:nnlm_ppl}, the perplexity is $80.89$ for the Subnet-$r_G$ (the full network) while $81.58$ for the full fixed model.

\begin{table}[h!]
    \centering
    \renewcommand{\arraystretch}{1.}
    \caption{Remaining percentage of computation ($C_t$), perplexity of NNLM on PTB w.r.t. the slice rate.}
    \resizebox{\columnwidth}{!}{
        \begin{tabular}{ c || cccccc|c}
    \thickhline
    Slice Rate $\mathbf{r}$ & 1.000 & 0.875  & 0.750  & 0.625  & 0.500  & 0.375  & 0.250  \\
    $\mathbf{C_t}$  & 100.0  & 76.56 & 56.25 & 39.06 & 25.00 & 14.06 & 6.250 \\
    \hline
    NNLM-1.0  & 81.58  & 85.23 & 91.04 & 99.68 & 116.5 & 155.5 & 298.8 \\
    NNLM-0.375  & \textbf{80.89}  & 81.79 & 82.86 & 84.65 & 87.92 & 91.17 & 112.1 \\
    NNLM-fixed  & 81.58  & \textbf{81.66} & \textbf{81.78} & \textbf{81.83} & \textbf{84.13} & \textbf{88.08} & 96.69 \\
    \thickhline
        \end{tabular}
    }
    \label{tab:nnlm_ppl}
\end{table}

This validates our hypothesis that the regularization and ensemble effect could improve the full model performance.
Further, the student-teacher knowledge distillation effect of the \textit{group residual learning} facilitates the learning process by transferring and sharing representation, and thus helps maintain the performance of subnets.


\subsection{CNNs for Image Classification}
\label{sec:exp_cnn}

\begin{table*}[ht]
    \centering
    \renewcommand{\arraystretch}{0.98}
    \caption{Configurations of representative convolutional neural networks on CIFAR (left panel) and ImageNet (right panel) datasets. Building blocks are denoted as ``[block, number of channels] $\times$ number of blocks''.
    }
    \resizebox{2.0\columnwidth}{!}{
        \begin{tabular}{ c||c|ccc||c|cc}
    
    \thickhline
    Group   &   Output Size &   VGG-13  &   ResNet-164  &   ResNet-56-2 &   Output Size & VGG-16 & ResNet-50    \\
    \hline
    conv1   &   32$\times$32  &   [conv3$\times$3, 64]$\times$2 &  [B-Block, 16]$\times$1    &   [B-Block, 16]$\times$1 &   112$\times$112 &  [conv3$\times$3, 64]$\times$3 & [B-Block, 64]$\times$1    \\
    conv2   &   32$\times$32  &  [conv3$\times$3, 128]$\times$2  &   [B-Block, 16]$\times$18    &   [B-Block, 16$\times2$]$\times6$&   56$\times$56  &  [conv3$\times$3, 128]$\times$3 & [B-Block, 64]$\times$3    \\
    conv3   &   16$\times$16  &    [conv3$\times$3, 256]$\times$2  &   [B-Block, 32]$\times$18    &   [B-Block, 32$\times2$]$\times6$&   28$\times$28 &  [conv3$\times$3, 256]$\times$3 & [B-Block, 128]$\times$4   \\
    conv4   &   8$\times$8  &  [conv3$\times$3, 512]$\times$4  &   [B-Block, 64]$\times$18    &   [B-Block, 64$\times2$]$\times6$&   14$\times$14       &  [conv3$\times$3, 512]$\times$3 &  [B-Block, 256]$\times$6 \\
    conv5   &   8$\times$8  &   -  &    -   &   -&   7$\times$7       &  [conv3$\times$3, 512]$\times$3 &  [B-Block, 512]$\times$3   \\
    avgPool/FC    &  10  &   [avg8$\times$8, 512] &   [avg8$\times$8, 64$\times$4] &   [avg8$\times$8, 64$\times$2$\times$4]&   1000    &    [512$\times$7$\times$7,4096,4096]    &    [avg7$\times$7,512$\times$4]    \\
    \hline
    Dataset & - & CIFAR & CIFAR & CIFAR & - & ImageNet-12 & ImageNet-12   \\
    Params & - & 9.42M & 1.72M & 2.35M & - & 138.36M & 25.56M \\
    \thickhline
        \end{tabular}
    }
    \label{tab:cnns_cifar}
\end{table*}

In this subsection, we evaluate \textit{\TopicWord} for convolutional layers on image classification tasks, mainly focusing on representative types of convolutional neural networks.
We first introduce dataset statistics for the evaluation.
Then configurations of the networks and training details are introduced.
Finally, we discuss and compare with baselines the results of \textit{\TopicWord} training scheme for CNNs.

\subsubsection{Datasets}
We evaluate the results on CIFAR~\cite{krizhevsky2009learning} and ImageNet-12~\cite{deng2009imagenet} image classification datasets.

The CIFAR~\cite{krizhevsky2009learning} datasets consist of $32 \times 32$ colors scenery images. 
CIFAR-10 consists of images drawn from 10 classes.
The training and testing sets contain $50,000$ and $10,000$ images respectively. 
Following the standard data augmentation scheme~\cite{he2016deep,huang2016deep,huang2017densely}, 
each image is first zero-padded with 4 pixels on each side, 
then randomly cropped to produce $32 \times 32$ images again, followed by a random horizontal flip.
We normalize the data using the channel means and standard deviations for data pre-processing.

The ILSVRC 2012 image classification dataset contains 1.2 million images for training and another 50,000 for validation from 1000 classes.
We adopt the same data augmentation scheme for training images following the convention\cite{he2016deep,zagoruyko2016wide,huang2017densely}, and apply a $224\times224$ center crop to images at test time.
The results are reported on the validation set following common practice.

\subsubsection{CNN Architectures and Training Details}
\label{sec:cnn_training_details}

\textit{\TopicWord[M][s]} dynamically slices channels within each layer in CNNs; thus we adopt three representative architectures differing mainly in the channel width for evaluation.
The first architecture is VGG~\cite{simonyan2014very} whose convolutional layer is a plain $3\times3$ \textit{conv} of medium channel width.
The second architecture is the pre-activation residual network~\cite{he2016identity} (ResNet).
ResNet is composed of the bottleneck block~\cite{he2016identity}, denoting as B-Block ($conv1\times1-conv3\times3-conv1\times1$).
We evaluate \textit{\TopicWord} on ResNet of varying depth and width, and denote the architecture adopted as ResNet-L, with L being the number of layers.
The third architecture is Wide Residual Network~\cite{zagoruyko2016wide}, which is denoted as ResNet-L-k, with k being the widening factor of the channel width for each layer.
Detailed configurations are summarized in Table~\ref{tab:cnns_cifar}.

To support \textit{\TopicWord}, convolutional layers and the batch-norm layers are replaced with counterpart layers supporting \textit{\TopicWord}.
For both baseline and \textit{\TopicWord} trained models, we train 300 epochs on CIFAR-10 with SGD of batch size 128 and initial learning rate 0.1, and 100 epochs on ImageNet-12 with SGD of batch size 128 and learning rate 0.01 with gradual warmup~\cite{he2016deep,goyal2017accurate}.
The learning rate is divided by 10 at 50\% and 75\% of the total training epochs for CIFAR-10, and at 30\%, 60\% and 90\% for ImageNet-12.
Other training details follow the conventions~\cite{he2016identity,zagoruyko2016wide}.

\begin{figure}[t]
    \centering
    \subfloat{\includegraphics[width=0.38\textwidth]{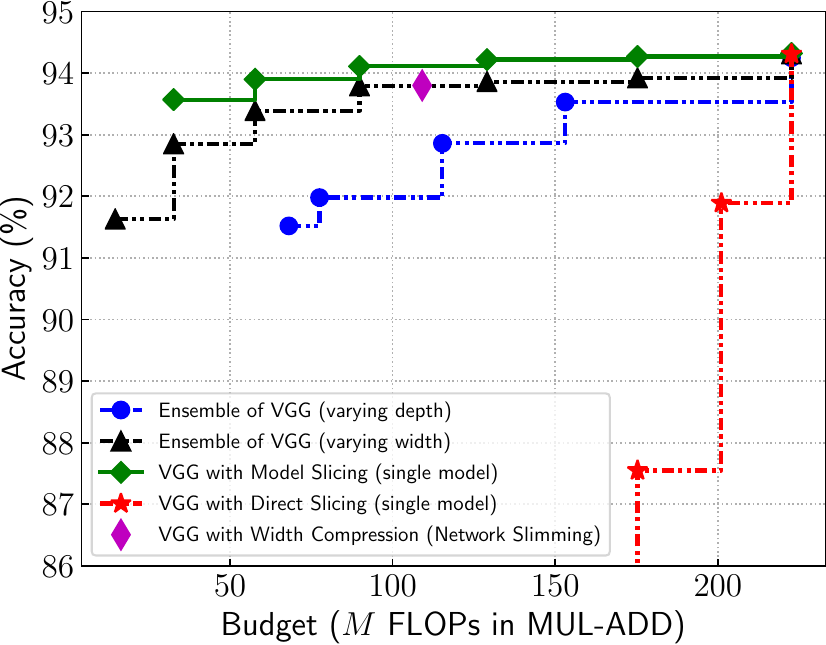} }
    \caption{Classification accuracy w.r.t. inference FLOPs of VGG-13 trained with model slicing against other baselines on the CIFAR-10 dataset.}
    \label{fig:cnn_comp}
\end{figure}

\subsubsection{Results of Model Slicing on CNNs}
\label{sec:cnn_results}
Results of representative CNNs on CIFAR and ImageNet datasets are illustrated in Figure~\ref{fig:resnet_flops}, Figure~\ref{fig:cnn_comp}, and summarized in Table~\ref{tab:res_percentage}.
In general, a CNN model trained with \textit{\TopicWord} is able to produce prediction with elastic inference cost by dynamically scheduling a corresponding subnet whose accuracy is comparable to or even higher than its conventionally trained counterpart.

\begin{table}[t]
    \centering
    \caption{Remaining estimated percentage of computation FLOPs ($C_t$)/parameter size ($M_t$), and accuracy of VGG-13, ResNet-164, ResNet-56-2 on CIFAR-10, and VGG-16, ResNet-50 on ImageNet w.r.t. the slice rate.}
    \resizebox{\columnwidth}{!}{
        \begin{tabular}{ c || ccccccc}
    \thickhline
    Slice Rate $\mathbf{r}$ & 1.000 & 0.8750 & 0.7500 & 0.6250 & 0.500 & 0.375 & 0.2500  \\
    $\mathbf{C_t}/\mathbf{M_t}$  & 100.0\% & 76.56\% & 56.25\% & 39.06\% & 25.00\% & 14.06\% & 6.25\% \\
    
    \hline
    VGG-13-lb-1.0  & 94.31 & 87.55 & 67.93 & 44.18 & 21.37 & 12.23 & \multicolumn{1}{|c}{10.19}\\
    VGG-13-fixed-models  & 94.31 & 93.92 & 93.86 & 93.79 & 93.39 & 92.85 & \multicolumn{1}{|c}{91.63}\\
    VGG-13-lb-0.375 & \textbf{94.32} & \textbf{94.27} & \textbf{94.22} & \textbf{94.11} & \textbf{93.90} & \textbf{93.57} & \multicolumn{1}{|c}{16.87}\\
    
    \hline
    ResNet-164-lb-1.0  & 94.96 & 87.55 & 67.93 & 44.12 & 21.37 & 12.33 & \multicolumn{1}{|c}{10.19}\\
    ResNet-164-fixed-models  & 94.96 & 94.85 & \textbf{94.68} & \textbf{94.35} & \textbf{94.13} & \textbf{93.65} & \multicolumn{1}{|c}{92.73}\\
    ResNet-164-lb-0.375 & \textbf{95.09} & \textbf{94.89} & 94.62 & 93.46 & 92.53 & 90.95 & \multicolumn{1}{|c}{16.83}\\
    \hline
    ResNet-56-2-fixed-models  & 95.25 & 95.20 & \textbf{95.17} & \textbf{95.01} & \textbf{94.52} & \textbf{94.04} & \multicolumn{1}{|c}{93.19}\\
    ResNet-56-2-lb-0.375 & \textbf{95.37} & \textbf{95.25} & 94.73 & 94.33 & 92.98 & 91.57 & \multicolumn{1}{|c}{10.58}\\
    \hline
    VGG-16-fixed-models  & 72.47 & - & \textbf{70.73} & - & 66.31 & - & 54.14 \\
    VGG-16-lb-0.25 & \textbf{72.53} & - & 70.69 & - & \textbf{66.41} & - & \textbf{54.20}\\
    \hline
    ResNet-50-fixed-models  & 76.05 & - & \textbf{74.73} & - & \textbf{72.02} & - & 63.91\\
    ResNet-50-lb-0.25 & \textbf{76.08} & - & 74.65 & - & 71.97 & - & \textbf{63.98}\\
    
    \thickhline
        \end{tabular}
    }
    \label{tab:res_percentage}
\end{table}

We compare the performance of \textit{\TopicWord[m][s]} with more baseline methods on ResNet in Figure~\ref{fig:resnet_flops}.
We can observe that ResNet-164 trained with \textit{\TopicWord[m][s]} (single model L164) achieves accuracies significantly higher than ResNet with Multi-Classifiers baseline, which confirms the superiority of \textit{\TopicWord} over depth slicing.
However, its performance is noticeably worse than the ensemble of ResNet of varying width, especially in the lower budget prediction.
This is mainly because the convolutional layer of ResNet-164 on CIFAR is narrow.
In particular, the convolutional layer in conv1/conv2 comprises 16 channels (see Table~\ref{tab:cnns_cifar}) and thus with \textit{slice rate} $0.375$, only 6 channels remain for inference which leads to limited representational power.
With twice the channel width, the single \textit{\TopicWord} trained model ResNet-L56-2 achieves accuracies comparable to the strong ensemble baseline of varying depth/width, model width compression baseline \textit{Network Slimming}~\cite{liu2017learning}, and achieves higher accuracies than SkipNet~\cite{wang2018skipnet} in corresponding inference budgets and generally better accuracy-budget trade-offs than MSDNet~\cite{huang2017multi}.
This demonstrates that \textit{\TopicWord} works more effectively for models of wider convolutional layers, e.g. the VGG-13, ResNet-L56-2 and ResNet-50.
For instance, the accuracy is 93.57\% for VGG-13-lb-0.375 with \textit{slice rate} 0.375, which is 0.72\% higher than its individually trained counterpart and takes around 14.06\% of the computation of the full network ($\sim$7.11x speedup).
This is also confirmed in the wider network VGG-16 and ResNet-50 on the larger dataset ImageNet.
Specifically, ResNet-50-lb-0.25 of \textit{slice rate} 0.25 achieves slightly higher accuracy than the fixed model of the same width and takes only around 6.25\% computation of the full network ($\sim$16x speedup).

We can also notice in Figure~\ref{fig:cnn_comp}, Table~\ref{tab:res_percentage} that the accuracy of CNNs trained conventionally (lower bound $lb$=1.0) decreases drastically as more channel groups are sliced off.
This shows that with conventional training, channel groups in the same convolutional layer are highly dependent on other groups in the representation learning such that slicing even one channel group off may impair the representation.
With the \textit{group residual representation learning} of \textit{\TopicWord}, one single network can achieve accuracy comparable to the ensemble of networks of varying width with significantly less memory and computational operation.

\subsection{Simulation of Cascade Ranking}
\label{sec:simulation}

We further simulate a cascade ranking scenario with six stages of classifiers.
CIFAR-10 test dataset is adopted for illustration which contains ten types of items (classes) and 1000 items (images) for each type, and VGG-13 (see Table~\ref{tab:cnns_cifar}) is adopted as the baseline model.
The classifier (model) is required to categorize each item into a type and then filter out all the items whose predicted category is not consistent with its previous type.
Therefore, the cascade ranking pipeline will only keep items of consistent classification type in \textit{all} the cascade models.
Typically, the pipeline deploys smaller models in the early stages to efficiently filter out irrelevant items, and larger but costlier models in subsequent stages for higher retrieval quality.
The baseline solution is a cascade model of the baseline model of varying width, which is compared with the \textit{\TopicWord} solution with corresponding sub-models sliced off the baseline model trained with \textit{\TopicWord}.
The parameter size and computation FLOPs of models at each stage are provided in Table~\ref{tab:cascade}.

\begin{table}[h]
    \centering
    \renewcommand{\arraystretch}{1.}
    \caption{Simulation of cascade ranking with the cascade model and the model trained with \textit{\TopicWord}.
    The \textit{precision} shows the prediction accuracy of each classifier;
    the \textit{aggregate recall} denotes the fraction of correctly classified items over the total number of items by each stage.
    }
    \resizebox{\columnwidth}{!}{
        \begin{tabular}{ c|c || cccccc}
    \thickhline
    \multicolumn{2}{c||}{\textbf{Stage/Classifier}} & 1st & 2nd & 3rd & 4th & 5th & 6th  \\
    \multicolumn{2}{c||}{\textbf{Model Width} ($\mathbf{r}$)} & 0.375 & 0.500 & 0.625  & 0.750  & 0.875 & 1.000  \\
    \hline
    \multicolumn{2}{c||}{\textbf{Params} (\textbf{M})}  & 1.33  & 2.36 & 3.68 & 5.30 & 7.21 & 9.42 \\
    \multicolumn{2}{c||}{\textbf{FLOPs} (\textbf{M})}  & 144.6 & 256.5 & 400.2 & 575.8 & 783.2 & 1022.5 \\
    \hline
    \multirow{2}{*}{\textit{\textbf{Cascade Model}}} & \textit{precision} & 92.85\% & 93.39\% & 93.79\% & 93.86\% & 93.92\% & 94.31\% \\
        & \textit{aggregate recall} & 92.85\% & 90.11\% & 88.62\% & 87.45\% & 86.70\% & 86.03\% \\
    \hline 
  \multirow{ 2}{*}{\textit{\textbf{\TopicWord[M][S]}}} & \textit{precision} & 93.57\% & 93.90\% & 94.11\% & 94.22\% & 94.27\% & 94.32\% \\
      & \textit{aggregate recall} & 93.57\% & 91.81\% & 89.47\% & 88.95\% & 88.76\% & 88.67\% \\
    \thickhline
        \end{tabular}
    }
    \label{tab:cascade}
\end{table}

Table~\ref{tab:cascade} summarizes the results on \textit{precision} and the \textit{aggregate recall} of each stage.
The results show two advantages of the \textit{\TopicWord} solution over the conventional cascade model solution: firstly, in terms of effectiveness, the \textit{\TopicWord} solution retrieves 88.67\% correct items in total as compared with 86.03\% of the conventional solution.
The significantly higher \textit{aggregate recall} is mainly because of the more consistent prediction between classifiers which we shall discuss and visualize in Section~\ref{sec:consistency};
secondly, in terms of efficiency, the conventional solution takes totally 29.3M parameters and 3182.8M FLOPs computation for the retrieval of each item, while \textit{\TopicWord} solution only takes 9.42M parameters in one model and the computation could be greatly reduced with the computation reusage discussed in Section~\ref{sec:residual_learning_effect}.

\subsection{Visualization}
\label{sec:visualization}

\subsubsection{Residual Learning Effect of Model Slicing}
\label{sec:residual_learning_vis}

\begin{figure}[t]
    \centering
    \subfloat[conv3]{\includegraphics[width=0.23\textwidth]{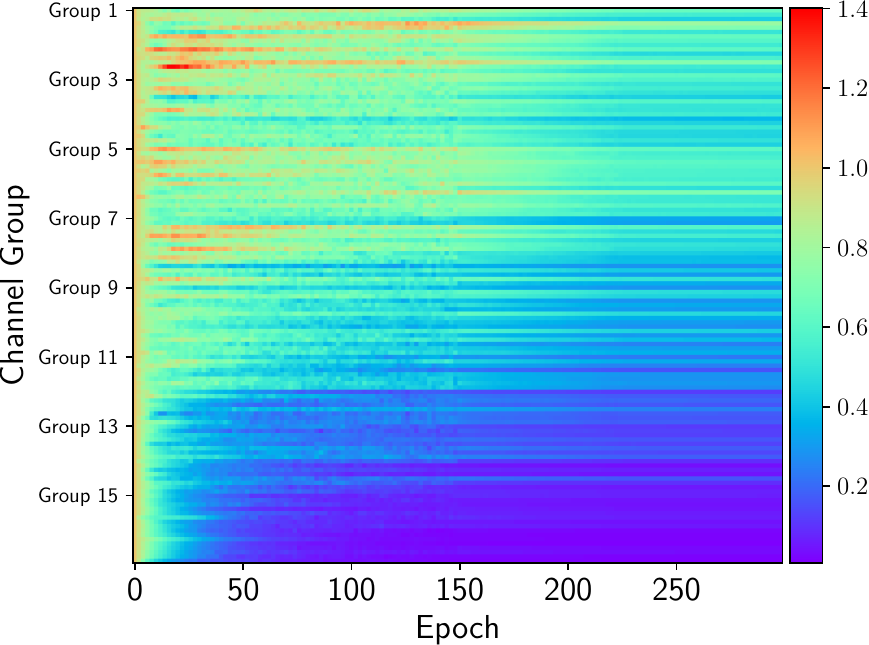} }
    \subfloat[conv5]{\includegraphics[width=0.23\textwidth]{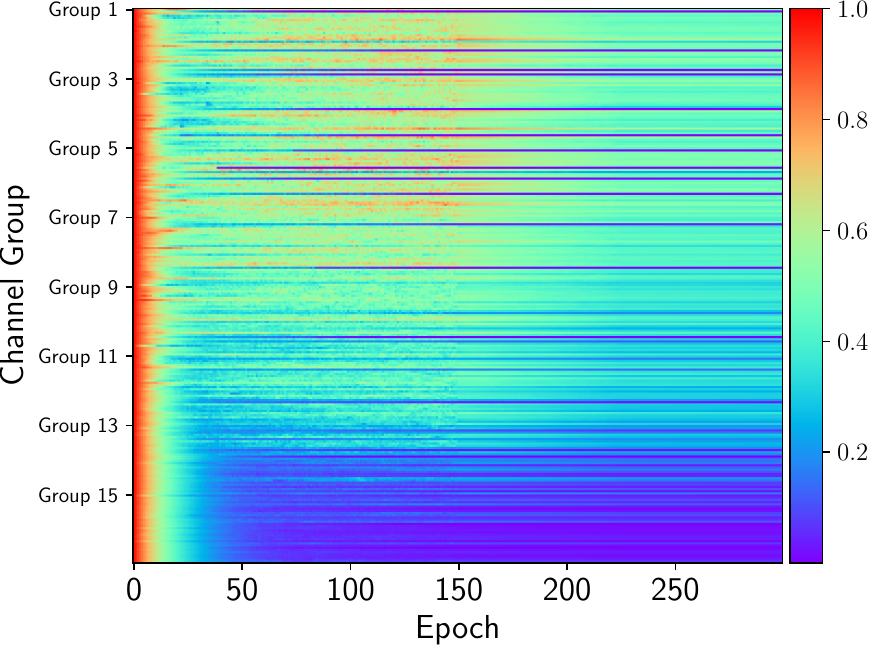} }
    \caption{Visualization of channel scaling factors ($\gamma$ from Equation~\ref{formular:batch_norm}) in scale as the training evolves, taken from the first convolutional layer of conv3, conv5 (Table~\ref{tab:cnns_cifar}) of VGG-13 trained on CIFAR-10 respectively. Brighter colors correspond to larger values.}
    \label{fig:vis_scale}
\end{figure}

In CNNs trained with \textit{\TopicWord}, each of the convolutional layers is followed by a group normalization layer to stabilize the scale of output with a scaling factor, i.e., $\gamma$ in Equation~\ref{formular:batch_norm}.
The scaling factor largely represents the importance of the corresponding channel.
We therefore visualize the evolution of these scaling factors during \textit{\TopicWord} training in Figure~\ref{fig:vis_scale}.
Specifically, we take the first convolutional layers of conv3 and conv5 in VGG-13 (see Table~\ref{tab:cnns_cifar}), which corresponds to low and high level feature extractors.
We can observe an obvious stratified pattern in Figure~\ref{fig:vis_scale}.
Groups from $G_1$ to $G_3$ of the base network gradually learn scaling factors of the largest values.
Meanwhile, from $G_3$ to $G_{8}$, the average scaling factor values gradually become smaller.
This validates our assumption that \textit{\TopicWord} training engenders \textit{residual group learning}, where the base network learns the fundamental representation and following groups residually build up the representation.

\subsubsection{Learning Curves of Model Slicing}

\begin{figure*}[t]
    \centering
    \subfloat[Test Error]{\includegraphics[width=0.36\textwidth]{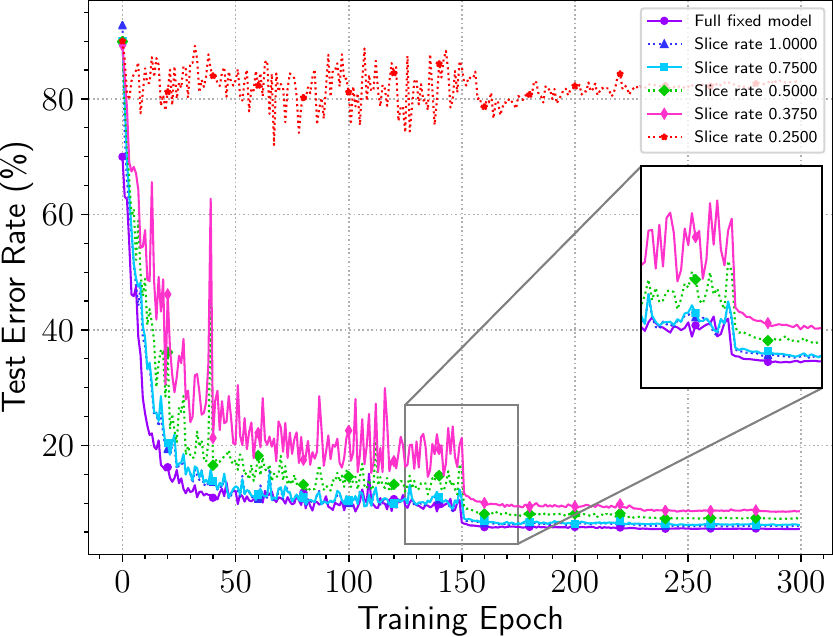} }
    \subfloat[Loss]{\includegraphics[width=0.36\textwidth]{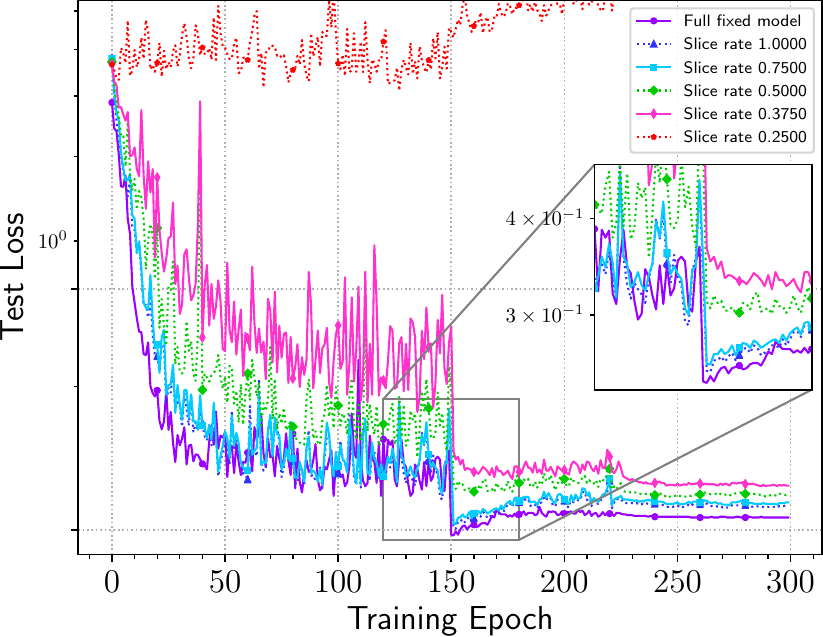} }
    \caption{Test Error Rate and Loss curves of VGG-13 full fixed model and VGG-13 trained with \textit{\TopicWord} ($r_1=0.375$) validated under different \textit{slice rates} on CIFAR-10 dataset.}
    \label{fig:vis_err_loss}
\end{figure*}

\begin{figure*}[t]
    \centering
    \subfloat[fixed models]{\includegraphics[width=0.36\textwidth]{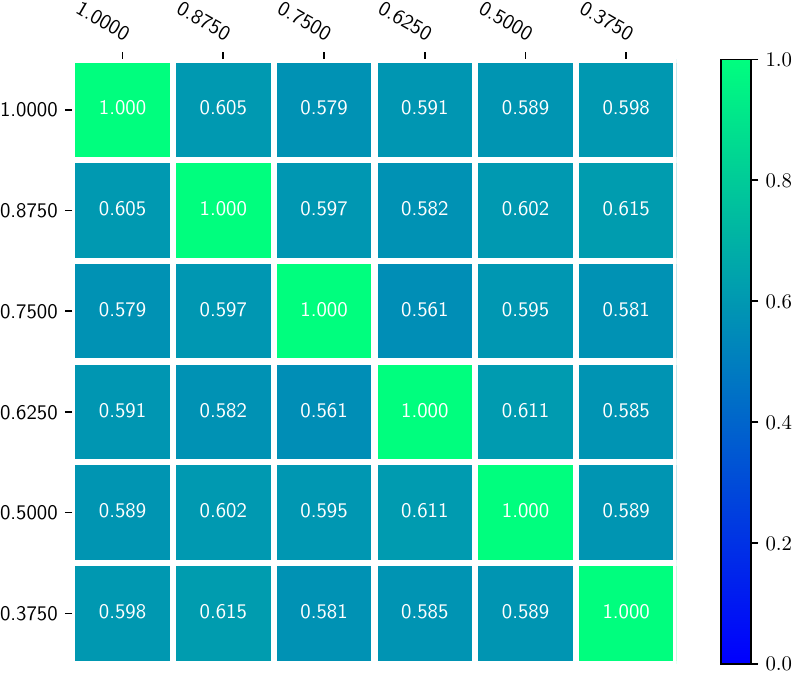} }
    \subfloat[subnets of the model trained with \textit{\TopicWord[m][s]}]{\includegraphics[width=0.36\textwidth]{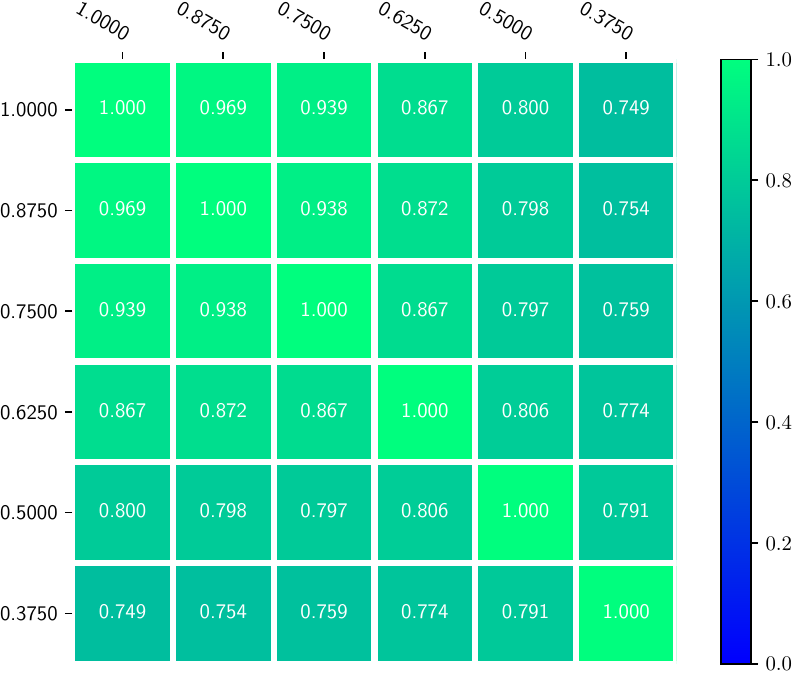} }
    \caption{Heatmap of the \textit{inclusion coefficient} of wrongly predicted samples between each pair of VGG-13 fixed models and sliced subnets of VGG-13 trained with \textit{\TopicWord} ($r_1=0.375$) respectively on CIFAR-10 dataset.}
    \label{fig:vis_heatmap}
\end{figure*}

Figure~\ref{fig:vis_err_loss} illustrates learning curves of VGG-13 trained with \textit{\TopicWord} compared with the full fixed model.
Learning curves of the subnets of VGG-13 trained with \textit{\TopicWord} reveal that the error rate drops faster in larger subnets and smaller subnets closely follow the larger subnets.
This demonstrates the knowledge distillation effect, where larger subnets learn faster and gradually transfer the knowledge learned to smaller subnets.
We notice that the final accuracy of subnets of a relatively larger slice rate approaches the full fixed model, which shows that the \textit{\TopicWord} trained model can trade off accuracy for efficiency by inference with a smaller subnet with less memory and computation at the cost of a minor accuracy decrease.

\subsubsection{Prediction Consistency of Model Slicing}
\label{sec:consistency}

We also evaluate the consistency of prediction results between the subnets of the model trained with \textit{\TopicWord}. 
Typically, the outputs are not the same for different models trained conventionally.
However, trained with \textit{\TopicWord}, the model of a larger slice rate incorporates models of lower slice rate as part of its representation.
Consequently, the subnets sliced off the \textit{\TopicWord} model are expected to produce similar predictions, and larger subnets could be able to correct wrong predictions of smaller models.
Figure~\ref{fig:vis_heatmap} shows the \textit{inclusion coefficient} of wrongly predicted samples between each pair of models. 
The inclusion coefficient measures the fraction of the wrongly predicted samples of the larger model over those of the smaller model.
It essentially measures the ratio of error overlapped between two models.
Unsurprisingly, the prediction results of \textit{\TopicWord} training is much more consistent than that of training different fixed models separately. 
Therefore, \textit{\TopicWord} may not be ideal for applications such as model ensemble which typically requires diversity, but could be extremely useful for applications requiring consistent prediction such as cascade ranking where the accumulated error is expected to be minimized.

\balance
\section{Conclusions}
\label{sec:conclusion}

Relatively few efforts have been devoted to neural networks dynamically providing predictions within memory and computational operation budget.
In this paper, we propose \textit{\TopicWord}, a general training framework supporting elastic inference cost for neural networks.
The key idea of \textit{\TopicWord} is to impose a structural constraint on basic components of each layer both during training and inference, and then regulate the width of the network with a single parameter \textit{slice rate} during inference given the resource budget on a per-input basis.
We have provided detailed analysis and discussion on training details of \textit{\TopicWord} and evaluated \textit{\TopicWord} through extensive experiments.

Results on NLP and vision tasks show that neural networks trained with \textit{\TopicWord} can effectively support on-demand workload by slicing a subnet from the trained network dynamically.
With \textit{\TopicWord}, neural networks can achieve significant reduction of run-time memory and computation with comparable performance, e.g., 16x speedup with \textit{slice rate} $0.25$.
Unlike conventional model compression methods where the computation reduction is limited, the required computation decreases quadratically to \textit{slice rate}.

\textit{\TopicWord[M]} also sheds light on the learning process of neural networks.
Networks trained with \textit{\TopicWord} engender \textit{group residual learning} in each layer, where components in the base network learn the fundamental representation while the following groups build up the representation residually.
Meanwhile, the learning process is reminiscent of knowledge distillation.
During training, larger subnets learn faster and gradually transfer the representation to smaller subnets.
Finally, \textit{\TopicWord} is readily applicable to the model compression scenario by deploying a proper subnet.

\section{Acknowledgments}



This research is supported by the National Research Foundation Singapore under its AI Singapore Programme [Award No. AISG-GC-2019-002] and Singapore Ministry of Education Academic Research Fund Tier 3 under MOE’s official grant number MOE2017-T3-1-007.




\newpage
\bibliographystyle{abbrv}
\bibliography{vldb}  









\end{document}